%% file: main.tex
\documentclass[conference]{IEEEtran}
\IEEEoverridecommandlockouts

\usepackage{amsmath,amssymb,amsfonts,amsthm,mathtools}
\usepackage[utf8]{inputenc}
\usepackage[T1]{fontenc}
\usepackage[hidelinks]{hyperref}
\hypersetup{pdfauthor={}, pdfsubject={}, pdfkeywords={}}
\usepackage{url}
\usepackage{booktabs}
\usepackage{microtype}
\usepackage{xcolor}
\usepackage{graphicx}
\usepackage{multirow}
\usepackage{cite}
\usepackage[capitalize,noabbrev]{cleveref}
\usepackage{flushend}

\usepackage{caption}
\input{math_commands}

\input{figures/TABLE_main_facts.tex}
\input{figures/TABLE_operating_point_facts.tex}

\input{figures/FIG_crosstask_facts.tex}

\title{Why Koopman Structure Does Not Help Differentiable World Models for Humanoid Whole-Body Tracking}

\author{\IEEEauthorblockN{Haizhou Ge$^{1}$, Mengshuo Wang$^{1}$, Yufei Jia$^{1}$, Lei Han$^{2}$, Lu Shi$^{1,\dagger}$, Guyue Zhou$^{1,\dagger}$, Ruqi Huang$^{1,\dagger}$}%
\thanks{$^{1}$Tsinghua University. $^{2}$DISCOVER Robotics.}%
\thanks{$^{\dagger}$Corresponding authors.}%
\thanks{\{ghz23, wangms25, jyf23\}@mails.tsinghua.edu.cn}%
\thanks{leo.han@discover-robotics.com, lus@nvidia.com}%
\thanks{zhouguyue@air.tsinghua.edu.cn, ruqihuang@sz.tsinghua.edu.cn}}

\begin{document}
\bstctlcite{IEEEexample:BSTcontrol}

\maketitle

\begin{abstract}
\input{sections/0_abstract}
\end{abstract}


\input{sections/1_introduction}
\input{sections/2_related_work}
\input{sections/3_method}
\input{sections/4_experiments}
\input{sections/5_conclusion}

\bibliographystyle{IEEEtran}
\bibliography{references}

\end{document}

%% file: math_commands.tex
\DeclareMathOperator{\diag}{diag}
\DeclareMathOperator{\clip}{clip}
\DeclareMathOperator{\LN}{LN}

\newcommand{\x}{\mathbf{x}}            
\newcommand{\xn}{\bar{\mathbf{x}}}     
\newcommand{\ctx}{\boldsymbol{\xi}}    
\newcommand{\ctxn}{\bar{\boldsymbol{\xi}}}
\newcommand{\act}{\mathbf{u}}          
\newcommand{\actn}{\bar{\mathbf{u}}}   
\newcommand{\inc}{\boldsymbol{\Delta}} 
\newcommand{\incn}{\bar{\boldsymbol{\Delta}}} 
\newcommand{\lift}{\mathbf{g}}         
\newcommand{\opt}{\boldsymbol{\mu}}    
\newcommand{\opthat}{\hat{\boldsymbol{\mu}}}
\newcommand{\noise}{\boldsymbol{\epsilon}}
\newcommand{\icpt}{\mathbf{a}}         
\newcommand{\gain}{M}                  
\newcommand{\bvec}{\mathbf{b}}         
\newcommand{\hist}{\mathbf{h}}         
\newcommand{\goal}{\mathbf{r}}         

%% file: figures/TABLE_main_facts.tex
\newcommand{\factSharedClips}{31}
\newcommand{\factReturnMlp}{0.830}

\newcommand{\factParamsMlp}{8.8}
\newcommand{\factReturnKoop}{0.823}

\newcommand{\factParamsKoop}{13.4}

\newcommand{\factLowerKoop}{21}

\newcommand{\factParamsAffine}{12.9}

\newcommand{\factReturnRankZero}{0.783}

\newcommand{\factParamsRankZero}{12.2}

\newcommand{\factParamsEncTwo}{9.2}

\newcommand{\factReturnBounded}{0.809}

\newcommand{\factReturnBoundedReliftEight}{0.780}

\newcommand{\factReturnLiftGrad}{0.823}

\newcommand{\factDeltaRankZeroVsKoop}{-0.020}
\newcommand{\factLowerRankZeroVsKoop}{30}
\newcommand{\factDeltaAffineVsKoop}{+0.0005}
\newcommand{\factLowerAffineVsKoop}{17}

\newcommand{\factDeltaBoundedVsKoop}{-0.003}
\newcommand{\factLowerBoundedVsKoop}{20}

\newcommand{\factReturnBoundedLiftOnce}{0.258}

\newcommand{\factIterBoundedLiftOnce}{4{,}000}
\newcommand{\factReturnBoundedAtLiftOnceEnd}{0.713}
\newcommand{\factReturnBoundedReliftEightAtLiftOnceEnd}{0.683}

\newcommand{\factRefSharedClips}{31}
\newcommand{\factRefFourHundredReturn}{0.815}
\newcommand{\factRefFourHundredFailed}{1/0/1}
\newcommand{\factRefFourHundredVsKoop}{-0.002}
\newcommand{\factRefFourHundredLowerVsKoop}{23}
\newcommand{\factRefHundredReturn}{0.802}
\newcommand{\factRefHundredFailed}{0/0/2}
\newcommand{\factRefHundredVsKoop}{-0.009}
\newcommand{\factRefHundredLowerVsKoop}{26}
\newcommand{\factRefFiftyReturn}{0.801}
\newcommand{\factRefFiftyFailed}{0/0/3}
\newcommand{\factRefFiftyVsKoop}{-0.012}
\newcommand{\factRefFiftyLowerVsKoop}{24}
\newcommand{\factRefTenReturn}{0.769}
\newcommand{\factRefTenFailed}{2/0/3}
\newcommand{\factRefTenVsKoop}{-0.033}
\newcommand{\factRefTenLowerVsKoop}{31}
\newcommand{\factRefOneReturn}{0.188}
\newcommand{\factRefOneFailed}{19/20/32}

%% file: figures/TABLE_operating_point_facts.tex
\newcommand{\factRecInputsGOneOneAffine}{0.71}

\newcommand{\factRecMuhatGOneOneAffine}{0.56}

\newcommand{\factRecInputsGOneOneKoop}{0.05}

\newcommand{\factRecMuhatGOneOneKoop}{0.58}

\newcommand{\factMlpErrGOneOneNone}{0.0909}
\newcommand{\factMlpErrGOneOneInputs}{0.1011}

\newcommand{\factRecInputsGOneFourAffine}{1.05}

\newcommand{\factRecInputsWindowsGOneFourAffine}{0.96}
\newcommand{\factRecInputsGOneFourKoop}{0.76}

\newcommand{\factRecInputsWindowsGOneFourKoop}{0.51}

\newcommand{\factRecInputsAntFourAffine}{-0.01}

\newcommand{\factRecInputsWindowsAntFourAffine}{0.01}
\newcommand{\factRecMuhatAntFourAffine}{1.01}
\newcommand{\factRecMuhatRatioAntFourAffine}{0.77}
\newcommand{\factRecInputsAntFourKoop}{0.07}

\newcommand{\factRecInputsWindowsAntFourKoop}{0.10}
\newcommand{\factRecMuhatAntFourKoop}{0.97}
\newcommand{\factRecMuhatRatioAntFourKoop}{0.75}
\newcommand{\factMlpErrAntFourNone}{0.0187}
\newcommand{\factMlpErrAntFourInputs}{0.0198}

\newcommand{\factMlpGainMuhatAntFour}{13}
\newcommand{\factMlpGainMuAntFour}{6}
\newcommand{\factGoalKoopNone}{1.24}
\newcommand{\factGoalKoopGoal}{1.38}
\newcommand{\factGoalKoopMu}{1.03}
\newcommand{\factTangentOpGOne}{1.00}
\newcommand{\factTangentMeanGOne}{31.2}

%% file: figures/FIG_crosstask_facts.tex
\newcommand{\factCrossGOneAffine}{1.54}

\newcommand{\factCrossGOneKoop}{1.31}

\newcommand{\factCrossGOneKoopMu}{1.09}

\newcommand{\factCrossHumanoidKoop}{1.55}

\newcommand{\factCrossHumanoidKoopMu}{1.29}

\newcommand{\factCrossAntKoop}{1.49}

\newcommand{\factCrossAntKoopMu}{1.13}

\newcommand{\factResidualAntOverGap}{22}

\newcommand{\factCrossWalkerKoop}{1.47}

\newcommand{\factCrossWalkerKoopMu}{1.36}

\newcommand{\factResidualWalkerOverGap}{67}

\newcommand{\factCrossCheetahKoop}{1.12}

\newcommand{\factCrossCheetahKoopMu}{1.08}

\newcommand{\factCrossHopperKoop}{0.96}

%% file: sections/0_abstract.tex
We test whether Koopman structure helps a differentiable world model used to train a simulated Unitree G1 whole-body tracking policy by backpropagation through imagined rollouts, replacing the MLP world model of an otherwise unchanged pipeline.
It does not: return is at best equal (0.823 versus 0.830), one-step error on identical data is 1.3 to 1.6 times the MLP's, and training is about 15\% slower.
Lifted rollouts diverge or fail to train policies; of the schemes tested, re-lifting at every imagined step trains best.
Re-lifting never uses the lifted linear dynamics and leaves a model affine in the action, with state-dependent offset and slope; an action-affine MLP also predicts worse, so the accuracy cost belongs to this class.
At a fixed state, the dynamics' curvature in the action explains only part of it.
Across states, the policy's actions span a curved region, so an affine model fits policy data only as the tangent at the policy's action, the operating point: on G1, the MLP's tangent there is as accurate as the MLP; at the mean action, 31 times worse.
The offset must encode, from the state alone, where the policy acts, whereas an MLP reads the action.
Supplying the policy's action removes most of the remaining gap on G1, Ant and Humanoid.
The only tested online repair approaching the MLP feeds the action into the nonlinear lift, abandoning the structure.
For this use, the MLP world model remains the better choice.

%% file: sections/1_introduction.tex
\section{Introduction}
\label{sec:intro}

A single policy can now track a whole library of human motions on a humanoid by backpropagating through rollouts imagined in a learned, differentiable world model, that is, by backpropagation through time (BPTT)~\cite{liao2026supermimic}, in the spirit of first-order policy learning through differentiable simulators~\cite{xu2022shac,suh2022differentiable}.
In this setting the world model is more than a predictor: the policy gradient is the derivative of its imagined rollouts, so its structure decides what the policy learns from.
Koopman models are an attractive structure for such a world model.
They lift the state into a space where the dynamics evolve linearly, which admits spectral constraints, closed-form adaptation and convex control~\cite{korda2016linear,bruder2020advantages,li2024continual,shi2024koopman}.

Koopman and bilinear latent models have been used to train policies in imagination~\cite{li2026koopman}, as features for a humanoid policy~\cite{jin2026dwmp}, for planning~\cite{pariente2026bilinear} and for long-horizon prediction~\cite{fathi2023course,mondal2023efficient}, but in settings that differ from ours (\cref{sec:related}), in which the world model predicts the robot's full state, root and joints, and the policy is trained by backpropagating through it.
To the best of our knowledge, no controlled comparison has asked whether the defining Koopman structure, a fixed lift with linear evolution and an action that enters only linearly or bilinearly, helps in this setting.
The closest evidence is that a control-affine model learns HalfCheetah dynamics worse than a full model in model-based RL, which Sung et al. attributed to the environment's complex nonlinearity, that is, to the model's representational capacity~\cite{sung2024robust}.

We answer this question on simulated Unitree G1 tracking of 40 LAFAN1 motion clips~\cite{harvey2020lafan}.
We replace the MLP world model of a SuperMimic-style tracker~\cite{liao2026supermimic} with a deep bilinear Koopman world model and keep the rest of the training pipeline and the evaluation unchanged; prediction accuracy is compared on identical data offline.
The answer is negative, and it has a structural explanation.
Theory argues against exact rollouts in the lifted space for systems with several limit sets~\cite{liu2023properties}, and the lifted rollouts of our Koopman world model do diverge; among the rollout schemes we tested, re-lifting the state at every imagined step trains the policy best.
Under re-lifting, the lifted linear dynamics that motivate the structure are never used across steps, and the predicted state increment becomes affine in the action, $\inc = \icpt(\x) + \gain(\x)\act$.
What remains is a restricted model class, whose extra error comes mostly not from the curvature of the true dynamics in the action but from the prediction at the \emph{operating point}, the noise-free action of the policy that collected the training data (the \emph{behavior policy}).
An action-affine model must in effect learn this operating point as a function of the state.

Our contributions are:
\begin{itemize}
\item We show that the Koopman world model gives no gain in return, dynamics adaptation, test-time model-predictive control or world-model prediction on held-out motions, while it predicts worse and trains slower than the MLP world model. An action-affine MLP also predicts worse, with no detected difference from Koopman's return, so the accuracy cost belongs to the action-affine class and not only to the Koopman parameterization.
\item We identify the operating point as the main source of the excess error beyond local non-affinity: supplying it removes most of that error. On G1 the state lacks the information to infer it; on a quadruped the data determine it, but the dynamics loss does not teach it.
\item We find that no online repair we tested both keeps the structure and closes the gap: lifts that keep the model action-affine give no gain; the only variant approaching the MLP, in prediction error and in return on the shared failure-free clips, feeds the action into the nonlinear lift.
\end{itemize}

%% file: sections/2_related_work.tex
\section{Related Work}
\label{sec:related}

\textbf{Koopman models with inputs.}
A Koopman model represents nonlinear dynamics by a linear operator acting on a lifted state~\cite{brunton2021modern,korda2016linear}.
With inputs, an exact finite-dimensional linear form exists only for systems whose nonlinearity is isolated from the control~\cite{shang2026existence}; control-affine systems admit bilinear lifts in continuous time~\cite{bruder2020advantages}, while discrete-time maps generally need input matrices that depend on both state and input~\cite{iacob2022koopman}.
Lifts may also read the input~\cite{proctor2016generalizing}, families of input-dependent operators cover general systems~\cite{haseli2023modeling,bevanda2024nonparametric}, and linear recurrences followed by nonlinear readouts are universal~\cite{orvieto2023universality}.
These results constrain what a fixed lift with bilinear inputs, the structure we study, can represent.

\textbf{Koopman and structured world models for control and learning.}
Koopman models work well where their linearity is used: convex model-predictive control of legged robots~\cite{li2024continual}, batched rollouts of open-loop action sequences~\cite{mondal2023efficient}, and long-horizon prediction with nonlinear action encoders~\cite{fathi2023course}.
Koopman Dreamer trains policies in imagination on DeepMind Control tasks with a spectrally constrained Koopman core and nonlinear stochastic corrections~\cite{li2026koopman}, Hao et al. estimate policy gradients from one-step Koopman predictions~\cite{hao2026efficient}, and DWMP uses a Koopman model of proprioception as an input feature for a humanoid policy~\cite{jin2026dwmp}.
Bilinear latent dynamics speed up planning~\cite{pariente2026bilinear}, locally linear latent models support iterative control~\cite{watter2015embed,levine2019prediction}, and physically structured models can improve over black-box ones~\cite{gupta2020structured,lutter2020differentiable}.

\textbf{Action-affine models and policy information.}
Sung et al., whose control-affine model failed to learn HalfCheetah dynamics in METRPO~\cite{kurutach2018metrpo}, instead linearized a full model at the policy's current action~\cite{sung2024robust}.
Control-affine models can be more accurate when identified from excitation data~\cite{xie2024learning} and are adopted for certifiable safety~\cite{zhou2026certifiable}.
On data collected by a policy, actions carry the policy's information, which confounds models conditioned on them~\cite{rezende2020causally,suau2023bad}, and conditioning a world model on an inferred policy representation improves its accuracy~\cite{chen2024policy}.
We connect these lines: in our setting, supplying the policy's action mainly helps the action-affine models, and most of their extra error lies at the operating point.

%% file: sections/3_method.tex
\section{Koopman World Models in BPTT Tracking}
\label{sec:method}

\subsection{Training Pipeline}
\label{sec:pipeline}

We build on SuperMimic~\cite{liao2026supermimic} and summarize the parts that matter here.
A single policy $\pi$ tracks a whole library of reference motions on the Unitree G1; its action $\act$ is a vector of joint position targets for PD control.
The policy reads a history $\hist$ of noisy observations and a tracking goal $\goal$ computed from the next reference frames.
While collecting data it acts with $\act = \opt + \noise$, where $\opt = \pi(\hist,\goal)$ is its noise-free action and $\noise$ is Gaussian exploration noise.
A world model, an MLP in SuperMimic, predicts the next state from the state features $\x$ (translation- and yaw-invariant root and joint quantities), the action and a simulation context $\ctx$ (friction, joint offsets, torso center-of-mass offset); it outputs an increment $\inc$ that is composed with the current state.
The world model is trained on replayed transitions with an eight-step rollout loss, and the policy is trained by backpropagating the tracking reward and a learned value through 32-step rollouts imagined in the world model.
Only the world model changes across the compared conditions; we call everything else the \emph{training pipeline}.

\subsection{The Koopman World Model}
\label{sec:koopman}

The Koopman world model lifts the normalized state and context, $\xn$ and $\ctxn$, and advances the lifted state linearly with a bilinear action term:
\begin{align}
\lift &= \big[\mathbf{o},\ \LN(\phi(\mathbf{o}))\big], \quad \mathbf{o} = \clip\big([\xn,\ctxn]\big), \label{eq:lift}\\
\lift' &= \lift + A\lift + \mathbf{c} + B\actn + H_{\mathrm{out}}\big((H_{\mathrm{s}}\lift)\odot(H_{\mathrm{in}}\actn)\big), \label{eq:step}\\
\incn &= C_1\lift + C_2(\lift' - \lift) + \mathbf{d}, \label{eq:decode}
\end{align}
where $\phi$ is an MLP encoder, $\clip$ saturates each normalized feature at $\pm 20$, $\LN$ is layer normalization and the bilinear term has low rank; $\actn = S(\act - \act_0)$ is the normalized action and $\incn$ the normalized increment, $\inc = D\incn + \mathbf{m}$, with diagonal scales $S, D$ and offsets $\act_0, \mathbf{m}$ from running statistics of the data.
Training adds a consistency loss that asks $\lift'$ to match the lift of the next state.
By \emph{Koopman structure} we mean the defining properties of this model: a fixed lift $\lift(\x,\ctx)$, linear evolution in the lifted space, and an action that enters only linearly or bilinearly.

The lift can be used in two ways.
\emph{Lifting once} lifts the first state and rolls the whole horizon out in the lifted space, decoding increments to integrate the physical state.
\emph{Re-lifting} lifts the integrated physical state again at every imagined step, which is the period-one case of periodic re-encoding~\cite{fathi2023course}.

\subsection{Re-Lifting Makes the Model Action-Affine}
\label{sec:affine}

Lifting once relies on the lifted linear dynamics across steps, and theory suggests that it cannot be exact here.
Liu et al. prove that a continuous-time system whose trajectories are precompact and which has more than one, but countably many, $\omega$-limit sets admits no continuous one-to-one immersion into a linear system~\cite{liu2023properties}.
Under a constant action, which makes the update~\eqref{eq:step} affine in $\lift$, the G1 comes to rest standing or in one of several fallen configurations, and the lift~\eqref{eq:lift}, which contains the clipped state, is one-to-one wherever no feature is clipped.
Whether the rest configurations are countable, and whether the result extends to the sampled map that the model learns, is not established for the G1, so we take the theorem as motivation and rely on the measured divergence (\cref{sec:help}).
Whole-body contact dynamics also do not isolate their nonlinearity from the control, which rules out an exact embedding that is linear in the input, though not by itself a bilinear one~\cite{shang2026existence}.
Re-lifting avoids the lifted rollout but reduces the model to a simple class.

Under re-lifting, $\lift$ is recomputed solely from $(\x,\ctx)$ at every imagined step, so substituting~\eqref{eq:step} into~\eqref{eq:decode}, with $\actn = S(\act - \act_0)$ and $\inc = D\incn + \mathbf{m}$, shows that the predicted increment is affine in the action at every fixed state and context, as a bilinear model is at a fixed lifted state~\cite{bruder2020advantages}:
\begin{equation}
\inc = \icpt(\x) + \gain(\x)\,\act,
\label{eq:affine}
\end{equation}
where, with the dependence on $\ctx$ left implicit,
\begin{align*}
\gain(\x) &= D\,C_2\big(B + H_{\mathrm{out}}\diag(H_{\mathrm{s}}\lift)\,H_{\mathrm{in}}\big)S,\\
\icpt(\x) &= D\big(C_1\lift + C_2(A\lift + \mathbf{c}) + \mathbf{d}\big) + \mathbf{m} - \gain(\x)\act_0.
\end{align*}

The policy is therefore trained through a world model of the class~\eqref{eq:affine}, which we call \emph{action-affine}, whereas an MLP world model $\hat{f}(\x,\act)$ can depend nonlinearly on the action.
The class is the control-affine form of Sung et al.~\cite{sung2024robust} with the policy's PD target, not a torque, as the input.

\textbf{Operating point, intercept and slope.}
To locate the error of a world model at a state $\x$, we perturb the behavior policy's noise-free action $\opt$ with Gaussian noise at the scale of the exploration noise and fit the predicted increments by least squares as
\begin{equation}
\inc \approx \bvec + \gain\,(\act - \opt).
\label{eq:decomp}
\end{equation}
We call $\opt$ the \emph{operating point}, and the fitted $\bvec$ and $\gain$ the \emph{intercept} and \emph{slope}.
Since the fit centers the actions on their sample mean, the intercept of an action-affine world model is its prediction at that mean, which equals $\opt$ up to sampling error: $\bvec \approx \icpt(\x) + \gain(\x)\opt$.
Its slope is exactly the action derivative $\gain(\x)$ that BPTT receives, and we call $\icpt(\x)$ its \emph{offset}, to tell it apart from the intercept. For the MLP world model and the simulator, the intercept and slope approximate the value at $\opt$ and the action Jacobian, because the dynamics are nearly affine at this scale (\cref{sec:alternatives}); the simulator's are the true intercept and slope, against which we compare each world model.

Policy data concentrate near the operating point, so to be accurate there an action-affine model has to match the tangent of the true one-step increment map $f(\x,\act)$ at $\opt$, which requires $\icpt(\x) \approx f(\x,\opt) - \gain(\x)\opt$.
When $f$ is curved in the action over the range the data cover, this tangent, and with it $\icpt(\x)$, depends on where the behavior policy acts, so the model must in effect learn the operating point as a function of the state through the dynamics loss.
Since $\opt = \pi(\hist,\goal)$ depends on the observation history and the tracking goal, $\x$ alone may lack the information, and where it suffices, the map from $\x$ to $\opt$ must still be learned.
An MLP evaluates $\hat{f}(\x,\act)$ at the actual action and needs neither.

%% file: sections/4_experiments.tex
\section{Experiments}
\label{sec:experiments}

\subsection{Setup}
\label{sec:setup}

\textbf{Online tracking.}
The task is tracking 40 LAFAN1 clips (2.45~h of motion)~\cite{harvey2020lafan} with a simulated Unitree G1 in MuJoCo Warp~\cite{todorov2012mujoco}, using 4{,}096 parallel environments and 12{,}000 training iterations; every other setting follows SuperMimic~\cite{liao2026supermimic}.
Each world model is trained with three seeds.
Evaluation runs each clip deterministically from its first frame to its end or to termination.
\emph{Return} is a clip's summed tracking reward divided by the maximum attainable, averaged over the 40 clips.
A clip \emph{fails} when its root height deviates from the reference by more than 0.25~m, or when the cosine of its root's tilt from the vertical differs from the reference's by more than 0.8 (for an upright reference, a tilt beyond about $78^\circ$).
Our primary measures are the 40-clip return and the failed clips of each run; paired per-clip comparisons additionally use only the clips that no compared run failed, a subset selected by outcome.
The online world-model error is the eight-step error on each run's own replay, so it differs across runs in data as well as in model.

\textbf{World models.}
The MLP world model has three hidden layers of width 2{,}048 (\factParamsMlp{}M parameters).
The Koopman world model uses a three-layer encoder of the same width, a 1{,}024-dimensional learned lift and a bilinear term of rank 512 (\factParamsKoop{}M).
In offline fits, wider lifts and higher ranks bring no appreciable gain, whereas a two-layer encoder, whose \factParamsEncTwo{}M parameters are close to the MLP's count, predicts worse (one-step error 0.121 against 0.110); online it also predicts worse than the three-layer encoder and earns less return than the MLP (\cref{tab:main}).
The \emph{affine MLP} outputs the offset and slope of~\eqref{eq:affine}, in normalized units, from a three-layer trunk of the same width (\factParamsAffine{}M), and the \emph{rank-0} Koopman world model drops the bilinear term (\factParamsRankZero{}M).
We call these \emph{parameterizations} of the action-affine class.
Unless named otherwise, \emph{the Koopman world model} is the model of \cref{sec:koopman}; its variants are named by what they change (rank 0, bounded, and the lifts of \cref{sec:repairs}).

\textbf{Offline fits.}
To compare accuracy on identical data, we fit the world models with the same architectures and eight-step loss to fixed datasets split into training and test parts by environment (G1) or by episode (Gymnasium), with 20{,}000 updates (40{,}000 in \cref{tab:operating_point}) and three fitting seeds unless stated otherwise.
The error is the mean squared one-step prediction error on the test part, with each increment dimension divided by its standard deviation in the training part, and we report it relative to the MLP fitted on the same data and seed.
Besides G1 data collected by the trained MLP-world-model policy with its exploration noise, we use the \texttt{medium-v0} Gymnasium MuJoCo datasets of Minari~\cite{towers2024gymnasium,younis2024minari}, about $10^6$ transitions each, collected by soft actor-critic and TQC agents~\cite{haarnoja2018sac,kuznetsov2020tqc}.
The \emph{non-affine residual} $E_{\mathrm{na}}$ of a dataset is measured in the simulator: we reset groups of environments to the same state, apply the behavior policy's actions perturbed as in the data, take one control step and fit an affine function of the action per state.
This residual lower-bounds the one-step error of any action-affine model on that action distribution.

\subsection{Does Koopman Structure Help? Is the Cost Koopman's?}
\label{sec:help}

\input{figures/TABLE_main.tex}

The Koopman world model earns no more return than the MLP world model (\cref{tab:main}): \factReturnKoop{} against \factReturnMlp{}, with seed ranges that overlap and a small, consistent per-clip deficit (lower on \factLowerKoop{} of \factSharedClips{} clips).
It is less accurate throughout late training, although its error is at times below the MLP's early on (\cref{fig:training}); on identical offline data, its one-step error is 1.3 to 1.6 times the MLP's, depending on the dataset (\cref{fig:crosstask}).
It has 52\% more parameters, and its training iterations are about 15\% slower in our implementation; size is not the only reason, since the two-layer encoder of \cref{sec:setup}, with about the MLP's parameter count, still takes 42\% longer per offline world-model update.
The structure thus adds cost without a measurable benefit.

\input{figures/FIG_training.tex}

The two ways of using the lift explain why re-lifting is the operative regime.
The lifted dynamics of the Koopman world model have spectral radius 1.94--2.23 across seeds, against about 1.08 for the physically unstable falling mode, and an eight-step rollout in the lifted space has normalized error of about 800--1{,}400.
A variant with a norm-bounded operator and a multi-step latent loss, the \emph{bounded} Koopman world model, rolls out 8--16 lifted steps.
Yet a policy trained through it by lifting once in imagination barely learns (\cref{tab:main}): at iteration \factIterBoundedLiftOnce{} it reaches \factReturnBoundedLiftOnce{}, against \factReturnBoundedReliftEightAtLiftOnceEnd{} when re-lifting every eight steps and \factReturnBoundedAtLiftOnceEnd{} when re-lifting every step, and at iteration 12{,}000 these two reach \factReturnBoundedReliftEight{} and \factReturnBounded{}.
Re-lifted every step, the bounded model predicts less accurately than the Koopman world model and earns slightly less return (per clip \factDeltaBoundedVsKoop{}, lower on \factLowerBoundedVsKoop{} of \factSharedClips{} clips).
Under re-lifting, the consistency loss constrains a property imagination never uses; zeroing its weight leaves the offline errors unchanged.

The affine MLP reproduces the pattern (\cref{tab:main}): it predicts worse than the MLP and somewhat worse than the Koopman world model, and its return shows no detected difference from Koopman's (\factDeltaAffineVsKoop{} per clip, lower on \factLowerAffineVsKoop{} clips).
The cost of Koopman structure in this setting is therefore the cost of the action-affine class, and among the parameterizations we tested the Koopman one is the most accurate offline on G1.
Restricting the class further does cost return: the rank-0 Koopman world model, whose slope is constant, roughly doubles the online error and loses return (\factReturnRankZero{}; per clip \factDeltaRankZeroVsKoop{} relative to the Koopman world model, lower on \factLowerRankZeroVsKoop{} of \factSharedClips{} clips).
On a common dataset, the relative slope error (the squared error of the slope in~\eqref{eq:decomp} over the squared magnitude of the true slope, both weighted by the perturbation covariance) is 0.032--0.040 for the Koopman world model and the affine MLP, 0.019 for the MLP and 0.152 for rank~0; return tracks it, but slope and total error rank the parameterizations identically, so these data do not separate the two.
This weak response of return to model error is consistent with the objective mismatch described by Lambert et al.~\cite{lambert2020objective}.

The structure brings no benefit in its other expected uses either.
We adapted both world models to three dynamics shifts (torso mass and principal inertias increased by 50\%, all motor gains reduced by 20\%, and both) with 128 to $8\times10^5$ shifted transitions and three seeds.
After adaptation, the MLP world model is more accurate than the Koopman one under each method that applies to both, from least-squares refits of the final linear layers to fine-tuning, and least-squares updates of the Koopman operator gain at most 6\% over refitting its decoder alone, and only with the most data.
Under both shifts combined, we also corrected the policy's actions at test time: at every control step, the world model imagines eight steps and five gradient steps adjust residuals added to the policy's actions.
With world models fine-tuned on about $2.6\times10^5$ shifted transitions, this model-predictive correction raises the return, normalized over the first 30~s of each clip, from about 0.69--0.72 to 0.84--0.85 with either the MLP or the bounded Koopman world model, while planning in the lifted space is 0.05 worse and not faster.
When clips are held out in five folds of eight and every world model is refitted without them (40{,}000 updates, one seed per fold), the eight-step errors of all world models on the held-out clips are 1.6 to 1.8 times those on seen clips, and on none of the 40 clips is the Koopman world model more accurate.
The one advantage we found is a property of the trained policy: without adaptation, the Koopman-trained policy is more robust when motors are weakened and the torso made heavier (0.541 against 0.412, higher on every seed), whereas under either shift alone the two policies' seed ranges overlap.

The weak coupling between error and return also holds when the model generates training data instead of gradients.
We ran model-based policy optimization~\cite{janner2019mbpo}, in which soft actor-critic~\cite{haarnoja2018sac} learns mainly from model rollouts, on Ant-v5 without contact-force observations.
The world model is an ensemble of five deterministic $4{\times}200$ networks, either all MLPs or all affine MLPs, and each ensemble type runs for 300{,}000 environment steps with ten seeds, in two batches of five.
Measured on the data that each run collected itself, the affine ensemble's one-step error is 2.1 times the MLP's, yet the final return shows no detectable cost (4{,}097 against 3{,}428).
The 90\% interval of the difference excludes a cost above about 9\% of the MLP's return if the ten seeds are treated as one sample, and above about 11\% when seeds are compared within batches.
The MLP baseline is below the return reported for this algorithm~\cite{janner2019mbpo}, and the setting differs from that of Sung et al.~\cite{sung2024robust}, so this result bounds the cost in this setting only.

\subsection{Is It Local Non-Affinity, Contact or Training?}
\label{sec:alternatives}

\input{figures/FIG_crosstask.tex}

Before turning to the operating point, we test the explanations that come to mind first: that the true dynamics are too curved in the action for any affine model, that the gap is peculiar to humanoids or to stiff contact, and that the action-affine models are undertrained or cannot identify their slope.
Only the first explains part of the gap.

\textbf{Local non-affinity.}
An action-affine model must fail where the true one-step map is curved in the action, but on policy data this explains only part of its excess error.
On G1, under the exploration noise of training, $E_{\mathrm{na}}$ is 2.4--2.7\% of the MLP's one-step error; under the wider spread of actions that the behavior policy takes at a given state, it is about a quarter of the Koopman--MLP gap.
On the Gymnasium datasets, $E_{\mathrm{na}}$ is \factResidualAntOverGap{}\% (Ant) to \factResidualWalkerOverGap{}\% (Walker2d) of the gap, and Walker2d's value is inflated because its observations clip velocities.

\textbf{Task and contact.}
The gap is not specific to humanoids (\cref{fig:crosstask}): the Koopman world model's error ratio is \factCrossHumanoidKoop{} on Humanoid, \factCrossAntKoop{} on Ant and \factCrossWalkerKoop{} on Walker2d; it is weak and variable on HalfCheetah (\factCrossCheetahKoop{}) and absent on Hopper (\factCrossHopperKoop{}).
On Walker2d and Hopper, the Koopman world model is not more accurate than the affine MLP, and over eight steps the action-affine models are more accurate than the MLP on HalfCheetah and Hopper (ratios 0.71 and 0.68 for the affine MLP, 0.87 and 0.74 for Koopman), so the mechanism below concerns one-step error.
Nor does the gap come from stiff contact or torque saturation: on G1 data recollected without torque limits, with softer contact, or with both, the one-step ratio stays between 1.47 and 1.57 (\cref{fig:mechanism}(a)).

\textbf{Training, data and exploration.}
The gap is not a shortfall of training: with 80{,}000 updates instead of 20{,}000 it widens to 1.69 on G1 and 2.27 on Humanoid, and on G1 the Koopman world model's test error rises after 40{,}000 updates while its training error keeps falling.
More data shrinks it on G1 (1.44 to 1.19 with four times the data, at equal training error) but not on Ant (1.67 to 1.74), where the action-affine models also fit the training data 1.7 to 1.9 times worse than the MLP.
Larger exploration noise, which would make the slope easier to identify, does not shrink the gap either: in single fits with up to 3$\times$ the noise on HalfCheetah and 2$\times$ on Ant, the gap stays or, on Ant, widens (1.49 to 1.66).
\Cref{sec:operating} traces most of the gap to where the behavior policy acts.

\subsection{The Operating Point}
\label{sec:operating}

\input{figures/TABLE_operating_point.tex}

On policy data, the action is almost the operating point: on G1, $\opt$ explains 99.7\% of the action variance.
When~\eqref{eq:decomp} is fitted to the online world models on G1 data from the MLP-world-model policy, about 90\% of the Koopman world model's excess one-step error over the MLP lies in the intercept, while its slope error is small.
The affine form itself is adequate where it is anchored: the seed-0 MLP linearized at $\opt$ is as accurate as the MLP on G1 (\factTangentOpGOne{} times its error), whereas the same MLP linearized at the mean action of the data has \factTangentMeanGOne{} times its error and the learned affine MLP (three seeds) \factCrossGOneAffine{} times.
The tangent at $\opt$ is also more accurate than the learned affine MLP on Ant, Humanoid and HalfCheetah, though not on Walker2d or Hopper.
This offers a compatible interpretation of the remedy of Sung et al.~\cite{sung2024robust}: a tangent anchored at the policy's current action need not learn the operating point from the state.

Supplying the operating point removes most of the gap that $E_{\mathrm{na}}$ does not explain on the high-dimensional tasks.
When every world model also reads $\opt$, the Koopman world model's ratio falls from \factCrossGOneKoop{} to \factCrossGOneKoopMu{} on G1, from \factCrossAntKoop{} to \factCrossAntKoopMu{} on Ant and from \factCrossHumanoidKoop{} to \factCrossHumanoidKoopMu{} on Humanoid, but only to \factCrossWalkerKoopMu{} on Walker2d and \factCrossCheetahKoopMu{} on HalfCheetah (\cref{fig:crosstask}).
On Ant, Humanoid and Walker2d, the gap that remains for the Koopman world model is 1.1 to 1.2 times $E_{\mathrm{na}}$.
The MLP responds much less (unchanged on G1 and Humanoid, about 7\% more accurate on Ant, with HalfCheetah an exception), as expected when the next state depends on the action actually taken and not on what the policy saw.

On G1, giving $\opt$ also supplies the goal and history information that $\x$ lacks.
\Cref{tab:operating_point} separates missing information from a failure to learn, on fresh collections (at 4$\times$, 2.87M training transitions and 1{,}024 test environments on G1, 3.50M and 500 on Ant).
We summarize each extra input $X$ by its \emph{recovered fraction} $(e_{\mathrm{none}} - e_X)/(e_{\mathrm{none}} - e_{\opt})$, where $e_X$ is the world model's 3-seed mean error given $X$, before division by the MLP's; it is 1 when $X$ helps as much as $\opt$ and can exceed 1.
We use these errors rather than ratios because the ratios in the table can fall merely because the MLP worsens, as it does on G1 with the policy's inputs (\factMlpErrGOneOneNone{} to \factMlpErrGOneOneInputs{} at 1$\times$).
On G1, the raw tracking goal does not help (in a separate collection, the Koopman ratio rises from \factGoalKoopNone{} to \factGoalKoopGoal{}, against \factGoalKoopMu{} with $\opt$), but the policy's own inputs, its goal encoding and observation history, recover most of the reduction once data suffice.
At 4$\times$ the recovered fraction is \factRecInputsGOneFourAffine{} for the affine MLP and \factRecInputsGOneFourKoop{} for the Koopman world model on every test transition (\factRecInputsWindowsGOneFourAffine{} and \factRecInputsWindowsGOneFourKoop{} on 16{,}384 sampled test windows), whereas at 1$\times$ it is \factRecInputsGOneOneAffine{} and \factRecInputsGOneOneKoop{}.
On G1 the operating-point cost is thus mainly missing information, and the Koopman world model needs more data to use it.

On Ant, the full observation that the policy reads recovers almost none of the reduction even at 4$\times$ (\factRecInputsAntFourAffine{} and \factRecInputsAntFourKoop{}; \factRecInputsWindowsAntFourAffine{} and \factRecInputsWindowsAntFourKoop{} on sampled windows), the MLP again worsens with it (\factMlpErrAntFourNone{} to \factMlpErrAntFourInputs{}), and the action-affine models fit the training data poorly as well.
We regress the data's actions on the full observation with a network of the world-model size (one seed, two-fold cross-fitting by environment, $R^2 = 0.98$ against $\opt$) and give the regression $\opthat$ to the world models.
At 4$\times$, $\opthat$ recovers \factRecMuhatAntFourAffine{} and \factRecMuhatAntFourKoop{} of the reduction in error, but it also makes the MLP \factMlpGainMuhatAntFour{}\% more accurate, more than $\opt$ does (\factMlpGainMuAntFour{}\%), so it carries more than the operating point; measured in ratios to the MLP given the same input, it recovers \factRecMuhatRatioAntFourAffine{} and \factRecMuhatRatioAntFourKoop{}.
On Ant, the dynamics loss thus does not teach an action-affine model the operating point, although the data determine it.
On G1, where the regression error is as large as the exploration noise, $\opthat$ recovers \factRecMuhatGOneOneAffine{} and \factRecMuhatGOneOneKoop{} at 1$\times$.

\input{figures/FIG_mechanism.tex}

Changing how the data's actions are chosen moves the gap, unlike changing the physics.
We recollect trajectories with a uniform random action replacing the policy's action with probability $p$.
As $p$ grows, the data's mean action, which takes the role of the operating point, varies less with the state, and the affine MLP's excess error beyond $E_{\mathrm{na}}$, a diagnostic rather than an exact decomposition, falls on HalfCheetah and Ant (\cref{fig:mechanism}(b)).
On HalfCheetah it falls from 0.12 to about zero, with a rise at $p = 0.5$; on Ant it falls from 0.51 to 0.18 at $p = 0.75$ but remains at 0.22 or 0.06 at $p = 1$, depending on how $E_{\mathrm{na}}$ is estimated.
Walker2d does not fall, but its $E_{\mathrm{na}}$ is unreliable with random actions.
The operating point is thus the main source of the extra error on policy data but not all of it.
Nor do our measures derived from it rank datasets by cost.
The difficulty of learning the tangent's offset from the state ranks the excess error of 26 Gymnasium datasets with Spearman correlation 0.39, comparable to that of learning $\opt$ itself.
The tangent at $\opt$ is within 1.5 times the MLP's error on only 9 of 27 datasets, none of them with random actions.

\subsection{Can a Repair Keep the Structure?}
\label{sec:repairs}

If the cost is the operating point, the direct repair is to let the lift read it, $\lift(\x,\ctx,\opt)$, with $\opt$ taken from the replay during model training.
With the gradient passed through $\opt$, the model approaches the MLP's online error and, on the shared clips, its per-clip return (\cref{tab:main}).
Its 40-clip return, however, equals the Koopman world model's (\factReturnLiftGrad{}) because one seed fails three clips.
In imagination, moreover, $\opt = \act$, so the action also enters the nonlinear encoder and the model is no longer action-affine.
Detaching $\opt$ yields the same forward map with a different derivative, and training collapses: all 40 clips fail by iteration 2{,}000.

A lift that reads a frozen copy of the policy, synchronized with the current policy every $N$ iterations, stays affine in the current action.
No synchronization interval improves on the Koopman world model, and more frequent synchronization is worse.
With $N = 400$, 100, 50 and 10, the 40-clip return is \factRefFourHundredReturn{}, \factRefHundredReturn{}, \factRefFiftyReturn{} and \factRefTenReturn{} (failed clips per seed \factRefFourHundredFailed{}, \factRefHundredFailed{}, \factRefFiftyFailed{} and \factRefTenFailed{}).
On the \factRefSharedClips{} clips that no run at these intervals, the MLP, the Koopman world model or the gradient variant failed, the per-clip return relative to the Koopman world model is \factRefFourHundredVsKoop{}, \factRefHundredVsKoop{}, \factRefFiftyVsKoop{} and \factRefTenVsKoop{} (lower on \factRefFourHundredLowerVsKoop{}, \factRefHundredLowerVsKoop{}, \factRefFiftyLowerVsKoop{} and \factRefTenLowerVsKoop{} clips).
With $N = 1$, training collapses (\factRefOneReturn{}; failed clips per seed \factRefOneFailed{}).
Among the online repairs we tested, the only one that approaches the MLP gives up the action-affine structure.

%% file: figures/TABLE_main.tex
\begin{table}[t]
\centering
\caption{Online G1 tracking at iteration 12{,}000. Return: mean $\pm$ s.d. over 3 seeds; failed: failed clips per seed; $\Delta$: seed-mean per-clip return minus the MLP world model's on the 31 clips that no run reaching iteration 12{,}000 failed (in parentheses: clips lower); error: online eight-step error with re-lifting at every step, averaged over the last 2{,}000 iterations. Middle: the bounded Koopman world model under three rollout schemes in imagination (\cref{sec:help}); lifting once is shown at iteration 4{,}000. Bottom: lifts that also read $\boldsymbol{\mu}$, with or without the gradient through it (\cref{sec:repairs}); the detached one is shown at its last checkpoint before it was stopped.}
\label{tab:main}
\footnotesize
\setlength{\tabcolsep}{2.5pt}
\begin{tabular}{@{}lcccc@{}}
\toprule
World model & Return & Failed & $\Delta$ & Error \\
\midrule
MLP & 0.830 $\pm$ 0.006 & 0/0/0 & --- & 0.101 \\
Koopman & 0.823 $\pm$ 0.006 & 0/2/0 & $-0.004$ (21) & 0.152 \\
Affine MLP & 0.821 $\pm$ 0.005 & 0/1/0 & $-0.004$ (21) & 0.165 \\
Koopman, rank 0 & 0.783 $\pm$ 0.014 & 1/2/3 & $-0.024$ (31) & 0.311 \\
Koopman, 2-layer encoder & 0.806 $\pm$ 0.012 & 1/1/2 & $-0.006$ (22) & 0.231 \\
\midrule
Re-lift every step & 0.809 $\pm$ 0.007 & 1/2/2 & $-0.007$ (27) & 0.203 \\
Re-lift every 8 steps & 0.780 $\pm$ 0.009 & 3/3/2 & $-0.009$ (27) & --- \\
Lift once & 0.258 $\pm$ 0.015 & 34/31/33 & --- & --- \\
\midrule
$\boldsymbol{\mu}$-lift, gradient & 0.823 $\pm$ 0.019 & 0/0/3 & $+0.004$ (11) & 0.107 \\
$\boldsymbol{\mu}$-lift, detached & 0.001 $\pm$ 0.000 & 40/40/40 & --- & --- \\
\bottomrule
\end{tabular}
\end{table}

%% file: figures/FIG_training.tex
\begin{figure}[t]
\centering
\includegraphics[width=\columnwidth]{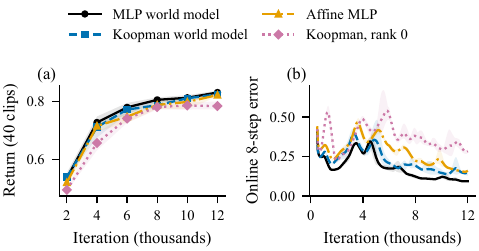}
\caption{Online G1 training, 3 seeds (mean and range): (a) return; (b) online 8-step error, moving
average over 500 iterations.}
\label{fig:training}
\end{figure}

%% file: figures/FIG_crosstask.tex
\begin{figure}[t]
\centering
\includegraphics[width=\columnwidth]{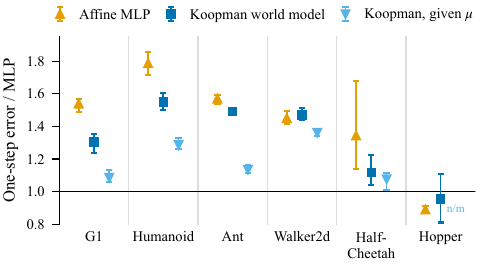}
\caption{Offline one-step test error relative to the MLP world model (marker: mean of the 3
per-seed ratios; whiskers: range). G1: a fresh collection of the tracking data; others: Gymnasium
MuJoCo datasets. Given $\boldsymbol{\mu}$: all world models, the MLP included, also read
$\boldsymbol{\mu}$; n/m: not measured.}
\label{fig:crosstask}
\end{figure}

%% file: figures/TABLE_operating_point.tex
\begin{table}[t]
\centering
\caption{One-step error on every test transition, as the 3-seed mean error of each world model over that of the MLP given the same extra input. Extra input: none; the policy's inputs; $\hat{\boldsymbol{\mu}}$, the data's actions regressed on them; or $\boldsymbol{\mu}$ (\cref{sec:operating}). Data: 4$\times$ uses the full training split of one collection per task and 1$\times$ a quarter of it; both use the same test set. ---: not measured.}
\label{tab:operating_point}
\footnotesize
\setlength{\tabcolsep}{3.5pt}
\begin{tabular}{@{}lllcccc@{}}
\toprule
Task & Data & World model & None & Inputs & $\hat{\boldsymbol{\mu}}$ & $\boldsymbol{\mu}$ \\
\midrule
G1 & 1$\times$ & Affine MLP & 1.62 & 1.34 & 1.50 & 1.40 \\
G1 & 1$\times$ & Koopman & 1.44 & 1.29 & 1.30 & 1.20 \\
G1 & 4$\times$ & Affine MLP & 1.37 & 1.14 & --- & 1.20 \\
G1 & 4$\times$ & Koopman & 1.19 & 1.01 & --- & 1.01 \\
\midrule
Ant & 1$\times$ & Affine MLP & 1.81 & 1.72 & 1.43 & 1.34 \\
Ant & 1$\times$ & Koopman & 1.67 & 1.56 & 1.35 & 1.26 \\
Ant & 4$\times$ & Affine MLP & 1.84 & 1.75 & 1.46 & 1.34 \\
Ant & 4$\times$ & Koopman & 1.74 & 1.61 & 1.35 & 1.22 \\
\bottomrule
\end{tabular}
\end{table}

%% file: figures/FIG_mechanism.tex
\begin{figure}[t]
\centering
\includegraphics[width=\columnwidth]{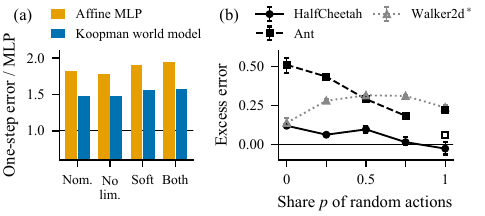}
\caption{(a) G1 data recollected by the same behavior policy under nominal physics (Nom.), without
torque limits (No lim.), with three times softer ground contact (Soft) and with both (one fit
each). (b) Excess error $(E_{\mathrm{aff}} - E_{\mathrm{MLP}} - E_{\mathrm{na}})/E_{\mathrm{MLP}}$ of
the affine MLP (3 seeds, mean and range). Ant's $E_{\mathrm{na}}$ for $p<1$ is measured at states
from the Minari dataset; at $p=1$ the filled marker uses states from the recollected data and the open one states from the Minari dataset. $^\ast$Walker2d's $E_{\mathrm{na}}$ is unreliable with random actions.}
\label{fig:mechanism}
\end{figure}

%% file: sections/5_conclusion.tex
\section{Conclusion}
\label{sec:conclusion}

In BPTT tracking on a simulated humanoid, Koopman structure does not help the differentiable world model.
Of the rollout schemes we tested, re-lifting at every step works best, and it reduces the Koopman world model to an action-affine world model whose lifted linear dynamics are never used across steps.
Action-affine world models predict worse than an MLP, with their extra error mostly in the prediction at the behavior policy's operating point rather than in the curvature of the dynamics; in our implementation the Koopman world model also trains more slowly, while return differs little between the MLP, Koopman and affine-MLP world models.
For training a tracking policy through a world model, the MLP world model therefore remains the better choice.

These conclusions concern one simulated robot and the Koopman structure of \cref{sec:koopman}; models whose action enters through a nonlinear encoder, or whose observables depend on the input, are outside it.
The link between the slope error and return is correlational, and the interventions that perturbed only the intercept or only the slope of an MLP world model in imagination collapsed training; interventions whose gradients stay consistent with the forward prediction could separate the two.
Two questions remain open.
One is why more frequent synchronization of the frozen reference policy degrades training.
The other is how a lift that reads the policy's inputs behaves online: it keeps the model action-affine and closed most of the gap offline on G1 with enough data, but its goal encoding would change with the policy during training.
Finally, Koopman structure may still pay off where its linearity is used, as in convex control of a low-dimensional subsystem or batched open-loop rollouts~\cite{li2024continual,mondal2023efficient}; weighing that benefit against the cost of the action-affine class measured here is left for future work.